\documentclass[
]{article}
\usepackage{amsmath,amssymb}
\usepackage{lmodern}
\usepackage{iftex}
\ifPDFTeX
  \usepackage[T1]{fontenc}
  \usepackage[utf8]{inputenc}
  \usepackage{textcomp} 
\else 
  \usepackage{unicode-math}
  \defaultfontfeatures{Scale=MatchLowercase}
  \defaultfontfeatures[\rmfamily]{Ligatures=TeX,Scale=1}
\fi
\IfFileExists{upquote.sty}{\usepackage{upquote}}{}
\IfFileExists{microtype.sty}{
  \usepackage[]{microtype}
  \UseMicrotypeSet[protrusion]{basicmath} 
}{}
\makeatletter
\@ifundefined{KOMAClassName}{
  \IfFileExists{parskip.sty}{%
    \usepackage{parskip}
  }{
    \setlength{\parindent}{0pt}
    \setlength{\parskip}{6pt plus 2pt minus 1pt}}
}{
  \KOMAoptions{parskip=half}}
\makeatother
\usepackage{xcolor}
\usepackage[normalem]{ulem}
\ifLuaTeX
  \usepackage{selnolig}  
\fi
\IfFileExists{bookmark.sty}{\usepackage{bookmark}}{\usepackage{hyperref}}
\IfFileExists{xurl.sty}{\usepackage{xurl}}{} 
\hypersetup{
  hidelinks,
  pdfcreator={LaTeX via pandoc}}

\author{}
\date{}

\begin{document}

\textbf{Mutagenesis screen to map the functions of parameters of Large
Language Models}

Yue Hu\textsuperscript{1}*, Gang Hu\textsuperscript{2}, Jixin Zheng\textsuperscript{3}, Patrick X.
Zhao\textsuperscript{1}, Ruimeng Wang\textsuperscript{4}*

1 Genetics Branch, NCI, NIH, Bethesda, MD, 20892, USA

2 Beijing Normal University, Beijing, 100875, China

3 Snap Inc., Santa Monica, CA, 90405, USA

4 Longevity Biomedical Inc., Bothell, WA, 98011, USA

* corresponding authors:

yuehu.mail@gmail.com

ruimemg@longevitybiomedical.com

\textbf{Abstract}:

\textbf{Large Language Models (LLMs) have markedly propelled
advancements in artificial intelligence, showcasing proficiency across a
wide range of tasks {[}1{]}. Despite a model\textquotesingle s
functionality being closely linked to its parameters, a systematic
approach to mapping these connections remains undeveloped. Models with
similar structures and parameter counts often perform differently
on various tasks, which has spurred detailed investigations into the
distinct patterns influencing their performance. We employed a
mutagenesis screening approach, drawing inspiration from biological
research methods, to probe into Llama2-7b and Zephyr. This method
involved altering elements within the models\textquotesingle{} matrices
to either their maximum or minimum values, thereby facilitating an
analysis of the interplay between model parameters and their
functionalities. Our study revealed intricate structures within both
models at multiple levels. While the phenotype map of many matrices
exhibited a mix of both maximum and minimum mutations, others displayed
a predominant sensitivity to one mutation type. Notably, mutations that
produced phenotypes, especially those with severe outcomes, tended to
cluster along axes. Furthermore, the placement of maximum and minimum
mutations typically followed a complementary pattern across the matrices
in both models, with the Gate matrix demonstrating a unique
two-dimensional asymmetry after rearrangement. In Zephyr, specific
mutations consistently yielded poetic or conversational outputs rather
than descriptive ones. These "writer" mutations were categorized by the
high-frequency initial word of the output, showing a significant
tendency to share the same row coordinate across different matrices. Our
mutagenesis screen has also uncovered many other significant variations
in the mutation maps of Llama2-7b and Zephyr, highlighting notable
discrepancies in their structural and functional attributes. Our
findings confirm that mutagenesis screening is a potent tool for
unraveling the complexities of LLMs and uncovering novel methods to
enhance their capabilities, thus offering profound insights into the
fundamental elements of AI systems.  Our code is publicly available at https://github.com/anonymous-submission200/Mutagenesis-Screen-Of-LLM.}

\textbf{Introduction}

Large Language Models (LLMs) represent one of the most significant
advancements in the field of artificial intelligence in recent years.
These models have revolutionized how we approach problems requiring
natural language understanding and generation, setting new benchmarks
for what machines can achieve in terms of linguistic and cognitive tasks
{[}1{]} . The exponential growth in the capabilities of LLMs can be
attributed to the increase in the number of their parameters, scaling
into the tens of billions or more. These parameters allow LLMs to
capture and generate nuances of human language with remarkable accuracy.
As a result, we\textquotesingle ve seen a surge in the emergence of
powerful functionalities that were previously thought to be the
exclusive domain of human intelligence. Despite their impressive
abilities, LLMs are often described as "black boxes" because the
intricate workings of their internal mechanisms are not fully
understood, even by the researchers who develop them.

To better understand how of LLMs operate, instead of simply taking them
as `black boxes', recent efforts have been engaged in explaining LLMs
from the local perspective. For example, Enguehard et al. {[}2{]}
utilized integrated gradient {[}3{]} to inspect the feature attribution
to specific outputs. Kobayashi et al. {[}4{]} proposed to track the
attention weights to analyze the linguistic capabilities of the
transformer blocks. Geva et al. {[}5{]} used on the feed forward network
(FFN) to show how FFN encodes human interpretable concepts. Despite all
these progresses, there is not a systematic method that can
comprehensively analyze the LLMs.

Drawing parallels with biological systems, which are similarly complex
and functionally diverse, this study adopts a biological
approach---specifically, mutagenesis screening---to probe into the
functionalities of LLMs. In biology, mutagenesis screening has been
instrumental in dissecting complex biological processes, from DNA
replication to protein transport, and decoding interactions within
cellular pathways {[}6,7{]}. By applying this method to the Llama2-7b
and Zephyr LLMs {[}8, 9{]}, we aim to uncover the underlying mechanisms
of these powerful models, enhancing our ability to refine and deploy
them more effectively.

\textbf{Mutation map statistics with different type of inputs}

Both Zephyr and Llama2-7b consist of 32 transformer layers. Each
transformer layer comprises one attention block, which includes four
matrices: K, Q, V, and O, and one MLP block containing three matrices:
Up, Down, and Gate {[}8,9{]}. In our research, we mutate the elements
within the attention matrices or MLP matrices of Llama2-7b or Zephyr to
study how the modifications affect the output of the model. We
implemented three distinct modification strategies on the matrix
elements: setting the element to zero, which we refer to as a zero
mutation, or replacing it with the maximum or minimum value found within
the matrix containing that element, henceforth termed maximum and
minimum mutations, respectively. In order to reduce the amount of
calculation, we simultaneously adjusted all elements within a 64x64
square section of the matrices. Each screen was performed with
comprehensive coverage across all matrix elements, without any overlaps
or missed elements. To follow biological terminology, we define the
output from a modified model that differs from the unaltered standard
model as a phenotype. A silent mutation is a mutation that does not
cause a phenotype, while a non-silent mutation (NSM) is one that does.
Our research primarily focuses on NSMs.

Our findings reveal that, regardless of the scenario, both the maximum
and minimum mutations cause a phenotype several orders of magnitude more
frequently than zero mutations in most cases (Supplementary Table 1).
Consequently, our subsequent analysis will concentrate on exploring the
effects of maximum and minimum NSMs only.

In our investigation into the NSMs map, we conducted a series of
experiments utilizing a diverse range of inputs: coding tasks involving
Python and Java for implementing bubble sort algorithms with random
seeds 0 and 10; descriptive tasks detailing the life cycle of
Drosophila, Newton's scientific contributions, and the signaling
pathways of p53; and a multiple-choice task with 21 selected questions
on international law from the Massive Multitask Language Understanding
(MMLU) dataset {[}10{]}, where the standard model provides correct
answers to the first 14 questions and incorrect answers to the remaining
7 questions. These experiments are referred to as Python0, Python10,
Java, Drosophila Newton, P53, and Mmlu21, respectively, and were
executed using one or both models (See Methods). Figure 1a displays the
total number of NSMs for each experiment, revealing that coding
experiments typically result in fewer NSMs than descriptive experiments.

The overlap analysis in Figure 1b for Llama2-7b reveals the greatest
mutation map similarity between the Python0 and Python10 experiments,
which use different random seeds. This is followed by the similarity
between Python and Java code requests that differ by a single word in
their input prompts. Comparing the mutation maps derived from
descriptive inputs with maps derived from code requests reveals even
less resemblance. The number of NSMs produced from the MMLU questions
and the descriptive inputs are comparable (Figure 1a). There is a
positive correlation in the number of NSMs per layer across different
experiments for Llama2-7b, suggesting a conservation of roles for
different layers across experiments (Figure 1c). Zephyr exhibits a
similar trend to Llama2-7b in terms of NSM production across different
inputs, except in the Mmlu21 experiment, where Zephyr demonstrates a
remarkable reduction in sensitivity to mutations (Figure 1a, c, Figure
2d). In the Mmlu21 experiment, more than one-third of matrices in Zephyr
harbor no NSMs (Supplementary Figure 1), whereas no other experiment in
either Llama2-7b or Zephyr has a single silent matrix lacking NSMs. Most
of these silent matrices in Zephyr\textquotesingle s Mmlu21 experiment
appear after layer 16, constituting two-thirds of all matrices after
this layer. All the attention or MLP matrices in layers 23 and 27 are
silent. The distinct response of Zephyr to the multiple-choice question
task, as opposed to other tasks where Llama2-7b remains consistent
across tasks, poses an intriguing question for further investigation.

Zephyr and Llama2-7b also shows significant differences in the
distribution of NSMs and phenotypes across layers for MLP matrices, as
illustrated in Figure 1e and detailed in Supplementary Table 1. It is
noteworthy that NSMs located in different parts of the matrix can yield
identical outputs, resulting in the same phenotype. The NSM number of
Llama2-7b MLP matrices has a wave-like distribution, with layer 0 and
layer 20-25 at the minima of the wave. The number of distinct phenotypes
of Llama2-7b of MLP matrices are significantly enriched in layer 30 and
31. In contrast, Zephyr displays a much more even distribution of both
NSMs and distinct phenotypes across layers. It is possible that this
distribution difference is correlated with the unsaturated training of
Llama2-7b as well as the better performance of Zephyr {[}4,5{]}.

In summary, our statistical analysis offers a comprehensive view of the
consistency and variations in mutation maps across different models,
experiments, layers, and matrices.

\textbf{Structure of NSM map}

In both Llama2-7b and Zephyr, certain matrices exhibit a pronounced bias
in sensitivity towards either maximum or minimum mutations. For example,
the NSM maps for the Down matrices in layers 2, 20, and 26 of Llama2-7b,
as well as layers 9, 15 and 22 of Zephyr, show a significant inclination
towards maximum NSM. Conversely, layers 17,18, 19 and layer 25 in
Llama2-7b, along with layers 13, 20, and 28 in Zephyr, exhibit a
preference for minimum NSM (refer to Figure 2 and Supplementary Figure
3). This sensitivity to either maximum or minimum NSM appears to be an
intrinsic characteristic of the models rather than a response to
specific types of input, indicating that it is a fundamental property of
the model itself. Notably, this bias is not consistently related to the
sequence of layers. For instance, Down matrices in layer 25 are more
sensitive to minimum mutation, whereas adjacent layers 24 and 26 are
more susceptible to maximum mutation. We find that 14\% of the matrices
in Llama2-7b \textquotesingle s NSM maps are biased towards maximum NSM,
and 19\% towards minimum NSM with a 20\% threshold of bias score (See
Method). In contrast, 14\% of matrices in Zephyr are biased towards
maximum NSM and 11\% towards minimum NSM. There appears to be no
consistent pattern of bias across different types of matrices within
layers for either model, nor is there a correlation in bias between
Llama2-7b and Zephyr across layers (Data not shown).

Secondly, in our analysis, we identified a pronounced enrichment of NSMs
in the column and row structures across all matrices in both models.
Furthermore, we pinpointed highly sensitive points within these
matrices, points at which mutations could trigger phenotypic changes
across a broad spectrum of inputs (Figure 3 and Supplementary Figure 1).
A standout observation was made in the V matrices of Llama2-7b, where
sensitive points across different layers coincided in specific column
positions --- column 22 (Supplementary Figure 1) and column 39 (Figure
3b, c). These particular points displayed a heightened sensitivity to
both maximum and minimum mutations, differentiated by a green color in
our visual representations, contrasting with the predominantly red
(maximum) or blue (minimum) NSMs observed elsewhere in the matrices. A
similar pattern of column-specific sensitivity was noted in the V
matrices column 32 across different layers of Zephyr, indicating a
consistent column where points were sensitive to both mutation types
across various inputs (Supplementary Figure 1). This pattern suggests an
underlying structural or functional significance to these specific
positions within the models, highlighting their critical role in the
models' responses to diverse inputs. Figure 3d reveals another type of
diversity in the mutagenesis study, showing the conservation of columns
in MLP Gate matrices across layers that was found in the Mmlu21
experiment but not in other experiments with Zephyr.

The third notable similarity involves a more nuanced structural aspect
observed in the NSM maps of many matrices of both Llama2-7b and Zephyr,
where distinct complementary patterns manifest along both x and y axes.
We refer to these patterns as \uline{C}orrelative C\uline{o}mplementary
\uline{Pa}tterns (COPA). In these patterns, if an element within two
complementary columns or rows is sensitive to maximum mutation, its
counterpart in the same row or column is likely sensitive to minimum
mutation, and vice versa (Figure 4a). To highlight COPA more clearly, we
selected for each matrix one row or column that demonstrates the best
overall correlation with all other rows or columns. We then rearranged
the rows or columns around this correlation vector, positioning
positively correlated rows or columns to the right in descending order
of correlation and negatively correlated ones to the left in ascending
order of correlation. Our analysis indicates that COPA can occur in any
attention and MLP matrices across any layer, though it is typically more
pronounced in the middle and later layers (Figure 4 a-c, Supplementary
Figure 2). Most matrices exhibit a relatively symmetrical structure
post-rearrangement, except for the Gate matrix, which shows a markedly
skewed structure during column reshooting (Figure 4c middle figure). The
reshoot along the x-axis displays a clear separation of maximum and
minimum NSMs, suggesting a significant directional bias on the x-axis
potentially related to the SiLU operation in the MLP computation of the
Gate input (Figure 4c lower figure). The intensity of COPA differs based
on the matrix and layer, as well as the specific experiment. The Python0
and Python10 experiments, conducted with Llama2-7b, have identical
conditions except for the use of different random seeds. Despite this,
the NSM maps from the Python10 experiment show noticeably stronger COPA
than those from the Python0 experiment, particularly in the MLP matrices
(Figure 4d, Supplementary Figure 2). Thus, unlike the patterns
previously discussed, COPA is a structure that is more sensitive to
variations.

In conclusion, our study identifies conserved features at the matrix,
columnar, row, and elemental levels in both Llama2-7b and Zephyr,
suggesting that similar characteristics are likely present in other
LLMs.

\textbf{The severity of phenotypes and their effect on function}

In our study illustrated in Figures 2-4, the colored dots indicate NSMs
that produce outputs differing from those of the standard models;
however, most of these NSMs cause only very weak phenotypes, such as a
single altered answer among 21 MMLU questions. We proceeded to examine
another critical attribute: the severity of phenotypes, measured by
performance scores on the MMLU or the degree of deviation from the
standard response for those descriptive experiments. This severity,
coupled with the specific impacts of NSMs, highlights the importance and
functional role of the affected sites within the model.

Our analysis initially focused on how NSMs affect performance on 21 MMLU
questions related to international law. Although many NSMs were benign,
generating typical "ABCD" responses with minimal effect on the overall
MMLU score, we identified two main types of detrimental NSMs. The first
type, "destructive NSMs", comprises a small subset that shifts responses
away from the standard "ABCD" format for some or all of the 21
questions. The second type, "underperforming NSMs", results in scores
below 10. Both types tend to cluster in specific columns within matrices
and some NSMs belong to both categories. (Figure 5a, Supplementary
Figure 3). Notably, underperforming NSMs with high scores tend to appear
in the same columns as those with low scores (Figure 5a, Supplementary
Figure 3). The V matrices for both Llama2b-7b and Zephyr are
particularly prone to these harmful mutations. In most V matrices from
the Mmlu21 experiment, column 39 in Llama2b-7b and column 32 in Zephyr
consistently contain the lowest-scoring NSMs (Supplementary Figure 3),
underscoring the significance of these structures and a consistency in
the mechanisms of both models. An exception is found in the Gate matrix
of layer 28 of Zephyr, where all the NSMs are destructive NSMs and align
along two rows rather than columns (Figure 5b). Among these destructive
NSMs, 34\% yield exclusively non-standard responses, while the remaining
66\% produce a mix of non-standard answers for some of the 21 MMLU
questions and standard "ABCD" responses for others. Notably, all the
standard responses generated by these destructive NSMs align with those
of the standard model, whether or not the standard
model\textquotesingle s answers are correct (Data not shown). This
pattern indicates that models containing some of these NSMs retain the
knowledge encoded in the standard model but struggle to deliver it in
MMLU task consistently.

For descriptive inputs, we assessed the severity of the mutations using
bag-of-words cosine similarity between the mutated and standard outputs,
assigning a cosine similarity score to each mutation\textquotesingle s
output. As with the MMLU test, NSMs characterized by larger deviations
tend to be concentrated along the same axes within each matrix.
Additionally, akin to the Mmlu21 experiment, for mutations with cosine
similarity below 0.5, those exhibiting higher cosine similarity are
often found in the same columnar or row structures as those with lower
similarity (Figure 5a, b).

The harmful NSMs show varied effects based on the evaluation
standard---either the MMLU test or descriptive outputs based on cosine
similarity. Some NSMs affect both standards, such as those in the
conserved columns of the V matrices of both models, or the rows in the
Gate matrix of layer 28 of Zephyr, highlighting their crucial role in
model functionality (Figure 5a, b). On the other hand, in
Zephyr\textquotesingle s layer 30 Down matrix, across all three
descriptive experiments---Drosophila, Newton, and P53---the harmful NSMs
that disrupt outputs and lead to low cosine similarity are consistently
found in multiple specific columns, highlighting the importance of these
columns in Zephyr for the descriptive task (Figure 5c). However, in the
Mmlu21 experiment, the model with these mutations produces answers
identical to the standard model, as there are no NSMs in Zephyr's layer
30 Down matrix (Figure 5c). Additionally, there are also mutations that
lower MMLU scores but do not affect the descriptive output of Drosophila
(Supplementary Table 2). This demonstrates the distinct responses of the
two evaluation methods to the same mutations, illustrating their
different sensitivities and underlying mechanisms in model assessment.
There are 421 NSMs from Llama2-7b and 922 NSMs from Zephyr that cause
severe disruptions in both the Drosophila experiment (cosine similarity
< 0.5) and the Mmlu21 experiment (score <= 8), concentrated
within only 21 matrices for Llama2-7b and 14 for Zephyr, respectively
(Figure 5d, e, Supplementary Figure 5). For both models, over 90\% of
NSMs are localized on columns within no more than 4 matrices. Figure 5d
and e depict some of these columns filled with green color dots in our
visual representations, indicating areas where the models are
particularly sensitive to modifications---whether increases or
decreases---in those specific elements. The severity of disruptions
noted in these columns and rows is consistent across all other
descriptive experiments when compared with Mmlu21 experiments
(Supplementary Figure 5). These sensitive locations, which significantly
affect model function when altered, are prime targets for further
investigation to understand the mechanisms of the models.

\textbf{Zephyr mutations with poetic or dramatic dialogues outputs}

In our investigation of the mutations' output of Drosophila experiments,
we identified rare mutations in Zephyr that produced outputs in the
forms of poems or dialogues.

The output from the standard Zephyr model is as follows (Supplementary
Table 3 row A):

\begin{quote}
\textless s\textgreater{} The life cycle of Drosophila:

The life cycle of Drosophila, commonly known as fruit flies, has four
distinct stages: egg, larva, pupa, and adult. Here is a detailed
explanation of each stage:

1. Egg stage: This stage begins when a female fruit fly lays an egg. The
eggs are laid in clusters, called egg masses, which are typically found
near the surface of rotting fruit or other organic material. The eggs
are oval-shaped and are translucent white in color. The eggs hatch into
larvae in approximately 24 hours.

2. Larva stage: This stage is also called the maggot stage.
\end{quote}

Here is one of the examples of the poetic outputs by certain mutations
(Supplementary Table 3 row D):

\begin{quote}
\textless s\textgreater{} The life cycle of Drosophila:

staring through a microscope

watching the embryos wriggle

a lesson in biology and life

these tiny creatures hold secrets

intricate patterns and genetic codes

a world inside a world

the first lesson: the egg

a yellow sphere, a potential universe

a future waiting to unfold

the second lesson: the embryo

a writhing mass of cells

a symphony of division and growth

the third lesson: the larva

a worm-like creature, a voracious eater

a race against time to pupate

the fourth lesson: the pupa

a cocoon,
\end{quote}

Another mutation produced a dialogue with intriguing twists
(Supplementary Table 3 row F):

\begin{quote}
\textless s\textgreater{} The life cycle of Drosophila:

otta: I\textquotesingle m thinking of becoming a fly.

dave: How\textquotesingle s that working out for you?

otta: (smiling) I\textquotesingle m still waiting to be born.

dave: Well, that could take a while. In the meantime, you might want to
get a little more specific. Are you thinking fruit fly or house fly?

otta: (confused) What\textquotesingle s the difference?

dave: Fruit flies are tiny and have a short lifespan. House flies are
bigger and live longer.

otta: (pauses) Then I think I\textquotesingle ll go for the fruit fly.
It sounds more challenging.
\end{quote}

While some poems and dialogues closely align with the intended topic of
the life cycle of Drosophila, others do not. To determine whether the
poetic and dramatic form of outputs were common or isolated, we fed the
model 151 different inputs, including descriptions, questions, and
factual statements (Supplementary Document 1). These mutations, which we
call "writer" mutations in the following, significantly increased the
production of poems, dialogues, and first-person narratives compared to
the standard model (see Supplementary Document 2). Notably, these
"writer" mutations consistently initiated outputs with different words
than the standard or other mutated models. For example, of the 151
outputs from the mutation in row D of Supplementary Table 3, 101 began
with "staring", in contrast to the standard Zephyr, which most
frequently began with "The" in 76 outputs---a commonality among many
NSMs. NSMs with "The" as the most frequent initial word, which make up
of more than 98\% of all NSMs, tend to have higher cosine similarity
values. In contrast, the remaining few \uline{R}are \uline{I}nitial
Words of \uline{H}ighest \uline{F}requency (RIHF) are more evenly spread
among NSMs with a cosine similarity of less than 0.8, as illustrated in
Figure 6a. These RIHF NSMs are not randomly distributed but show a
constrained pattern. Of 224 attention and MLP matrices in total, they
are notably concentrated in just 21 of matrices, particularly enriched
in the Down matrix of layers 1 and 30 (Figures 6b and Supplementary
Figure 6). NSMs within a single RIHF group share identical or a limited
number of y-coordinates even when they are on different matrices and
different layers (Figures 6c and 6d), while NSMs across different RIHFs
often share x-coordinates, results in a structured enrichment within
specific matrix columns (Figures 6d and Supplementary Figure 6). This
pattern significantly overlaps with maps of low cosine similarity NSMs,
particularly in the Down matrix of layer 30 where there is a perfect
correlation with the columnar patterns that were conserved across all
descriptive tasks of Zephyr (Figures 6d and 5c).

.

The quality of output from "writer" mutations correlates strongly with
their Mmlu21 scores. For instance, mutations with the RIHF "staring"
produce the highest quality outputs when their Mmlu21 answers are
identical to those of the standard model. Outputs from mutations with
answers close to but not matching the standard model perfectly are of
lesser quality, and those with low Mmlu21 scores struggle to produce
coherent outputs (see Supplementary Table 3, rows A-G, and Supplementary
Document 2). Despite notable changes in output style, mutations from the
first two groups retain a high level of relevance and performance,
demonstrating excellent comprehension across most of the 151 inputs in
our study (Supplementary Document 2). They perform comparably on Python0
and Java tasks and achieve similar scores on a complete MMLU test (Data
not shown). Among the 151 inputs, the last 50 are factual statements
while the first 101 are more literary expressions (Supplementary
Document 1). We observed that "writer" mutations generally excel in the
first 101 questions, producing elegantly styled poems and dialogues with
interesting twists (Supplementary Table 3, column "The life cycle of
Drosophila"), but their performance noticeably declines in the factual
statements compared to the standard model (Supplementary Document 2).
RIHF NSMs are also present in Llama2-7b and follow the same distribution
rules (Figures 6A-C), although no change in output form was observed in
these NSMs as in Zephyr.

To summarize, here we identified intriguing "writer" mutations in Zephyr
that show a strong tendency to produce poetic or dialogic outputs. These
"writer" mutations belong to a broader group with RIHF, which are
subject to highly restrictive spatial constraints in the matrices. These
RIHFs are localized predominantly within specific columns of several
matrices, particularly in select Down matrices. Notably, RIHFs with the
same initial words mostly sharing a single or very few row coordinates,
regardless of their matrix location. (Figure 6c, d).

\textbf{Conclusion}

This study has successfully utilized a mutagenesis screening approach
inspired by biological research methods to explore the complexities of
LLMs, specifically Llama2-7b and Zephyr. This method reveals
multifaceted features across both models. At the matrix level, certain
matrices in both models displayed predominant sensitivity to either
maximum or minimum mutations. At the columnar and row levels, NSMs,
particularly those significantly influencing MMLU performance or
altering model responses, notably clustered along axes. At the element
level, we observed COPA phenomenon, which impose strict distribution
constraints on NSMs, and identified unique biases in the Gate matrix not
present in other matrix types.

The similarity in parameter counts and structural design between
Llama2-7b and Zephyr has provided a solid foundation for employing
mutagenesis screening to study correlations between matrix parameters
and model performance. Notably, the MLP matrix exhibited an uneven
distribution of NSMs and a marked concentration of distinct phenotypes
in the last two layers of Llama2-7b, suggesting that in comparison with
Zephyr there are more parameters not fully utilized by Llama2b-7b, and
the last two layers of Llama2b-7b bear a disproportionate burden in
realizing the model\textquotesingle s potential, which probably
underlies the underperformance of Llama2b-7b comparing with Zephyr
{[}5{]}. An interesting divergence was observed in how Zephyr exhibited
striking stability in the Mmlu21 experiment and had many fewer NSMs in
this task, while Llama2-7b showed comparable NSM statistics across
various tasks. This distinction in task-specific performance between the
models warrants further exploration.

Importantly, the discovery of "writer" mutations in Zephyr, which
generate creative and thematic outputs such as poetry and dialogues,
opens new avenues for utilizing LLMs beyond traditional tasks. These
mutations not only highlight the models\textquotesingle{} capacity for
generating inventive content but also suggest that such features can be
further developed and harnessed. It also raises intriguing questions
about the consistency of row coordinates across layers for NSMs with the
same RIHF, like those showing "staring" consistently across specific
layers.

Looking forward, the mutagenesis screening approach has considerable
potential for expansion. For instance, although the use of basic cosine
similarity metrics in this study proved to be feasible, this approach
makes use of the minimum information from the output, and thus points to
the need for more sophisticated analytical methods to fully leverage the
rich data generated by mutagenesis screening. Moreover, identifying more
mutations that produce novel outputs, delving deeper into the functions
of specific elements, or generating function maps of complex tasks like
context understanding, emotion recognition, and grammatical correction
could reveal crucial insights into our understanding of how LLMs
process. In addition, our method is highly adaptable and can be applied
to various other large-scale models, such as image and video generative
models {[}11-14{]}, and multi-modal LLMs {[}15-18{]}, to provide
profound research insights in diverse aspects of artificial
intelligence.

In conclusion, the success of this study in applying biological research
methods to the study of artificial intelligence not only enhances our
understanding of LLMs but also sets the stage for future advancements in
AI design and application. The integration of methodologies from
biological research into AI studies promises a rich reservoir of
techniques that can propel our understanding and innovation in
artificial intelligence forward. By bridging the gap between biological
and AI research, we can enrich our toolkit for exploring both the
artificial \textquotesingle brains\textquotesingle{} of LLMs and the
biological neural networks, potentially catalyzing breakthroughs across
both fields.

\textbf{Reference}

{[}1{]} Naveed H et al. ``A Comprehensive Overview of Large Language
Models'' In arXiv preprint arXiv:\textbf{2307}.06435, 2024.

{[}2{]} Enguehard J. ``Sequential Integrated Gradients: a simple but
effective method for explaining language models'' In arXiv preprint
arXiv:\textbf{2305}.15853, 2023.

{[}3{]} Sundararajan M, Taly A, Yan Q. ``Axiomatic attribution for deep
networks'' in \emph{International conference on machine learning. PMLR},
2017: 3319-3328.

{[}4{]} Kobayashi G, Kuribayashi T, Yokoi S, et al. ``Analyzing
feed-forward blocks in transformers through the lens of attention map''
In arXiv preprint arXiv:\textbf{2302}.00456, 2023.

{[}5{]} Geva M, Caciularu A, Wang K R, et al. ``Transformer feed-forward
layers build predictions by promoting concepts in the vocabulary space''
In arXiv preprint arXiv:\textbf{2203}.14680, 2022.

{[}6{]} Bökel C ``EMS screens : from mutagenesis to screening and
mapping'' In: \emph{Methods Mol Biol} \textbf{420}, 2008.

{[}7{]} Przybyla L and Gilbert LA.``A new era in functional genomics
screens'' In \emph{Nat Rev Genet} \textbf{3} 23:89, 2022.

{[}8{]} Touvron H et al. ``Llama 2: Open Foundation and Fine-Tuned Chat
Models'' arXiv preprint arXiv:\textbf{2310} 16944, 2023.

{[}9{]} Tunstall Let al. ``Zephyr: direct distillation of LM alignment''
Performance arXiv preprint arXiv:\textbf{2307} 09288, 2023.

{[}10{]} Hendrycks, D., Burns, C., Basart, S., Zou, A., Mazeika, M.,
Song, D., and Steinhardt, J. Measuring massive multitask language
understanding. arXiv preprint arXiv:\textbf{2009}.03300, 2020.

{[}11{]} Rombach R, Blattmann A, Lorenz D, et al. ``High-resolution
image synthesis with latent diffusion models{[}`` In \emph{Proceedings
of the IEEE/CVF conference on computer vision and pattern recognition}.
2022: 10684-10695.

{[}12{]} Blattmann A, Dockhorn T, Kulal S, et al. ``Stable video
diffusion: Scaling latent video diffusion models to large datasets'' In
arXiv preprint arXiv:\textbf{2311}.15127, 2023.

{[}13{]} Yu L, Cheng Y, Sohn K, et al. ``Magvit: Masked generative video
transformer'' In \emph{Proceedings of the IEEE/CVF Conference on
Computer Vision and Pattern Recognition}. 2023: 10459-10469.

{[}14{]} Yu L, Lezama J, Gundavarapu N B, et al. ``Language Model Beats
Diffusion-\/-Tokenizer is Key to Visual Generation'' In arXiv preprint
arXiv:\textbf{2310}.05737, 2023..

{[}15{]} Liu H, Li C, Wu Q, et al. ``Visual instruction tuning'' In
\emph{Advances in neural information processing systems}, 2024, 36.

{[}16{]} Bai J, Bai S, Yang S, et al. ``Qwen-vl: A frontier large
vision-language model with versatile abilities'' In arXiv preprint
arXiv:\textbf{2308}.12966, 2023.

{[}17{]} Chen Z, Wu J, Wang W, et al. ``Internvl: Scaling up vision
foundation models and aligning for generic visual-linguistic tasks'' In
\emph{Proceedings of the IEEE/CVF Conference on Computer Vision and
Pattern Recognition}. \textbf{2024}: 24185-24198.

{[}18{]} Wang W, Lv Q, Yu W, et al. ``Cogvlm: Visual expert for
pretrained language models'' In arXiv preprint
arXiv:\textbf{2311}.03079, 2023.

\textbf{Figure Legends}

\textbf{Figure 1: Comparative Mutation and Phenotype Analysis of
Llama2-7b and Zephyr}

\textbf{(a) Total counts of non-silent mutations (NSMs) across all
experiments.}

\textbf{(b) Mutation map overlaps across experiments.} The overlap
calculation includes both silent and non-silent mutations. See Method
section for detail.

\textbf{(c)} \textbf{Comparative NSM analysis across experiments.} Both
x and y coordinates represent the number of NSMs per layer for a given
experiment. For Llama2-7b (Left) and Zephyr (Middle), the x-coordinate
is based on the Python0 experiment, while the y-coordinates are derived
from other experiments. Additionally, for both Llama2-7b and Zephyr
(Right), the x-coordinate is based on the Python0 experiment, with the
y-coordinate representing the Mmlu21 experiment. Notably, there is a
positive correlation in the number of NSMs per layer across different
experiments, with the exception of the Mnlu21 experiment for Zephyr.

\textbf{(d)} \textbf{Counts of NSMs} \textbf{for each attention and MLP
matrix of Python0 experiments.}

\textbf{(e) Layer-specific NSM (blue) and distinct phenotype (green)
counts in Python0 experiments for the Gate, Up, and Down matrices.} The
x-coordinate represents the ordered rank of each layer. Llama2-7b shows
a more uneven distribution of both NSM and distinct phenotypes across
layers compared to Zephyr.

\textbf{Figure 2: Imbalance in Maximum and Minimum NSMs Across Selected
Matrices}

\textbf{(a) Example of bias towards maximum NSM in Llama2-7b: Attention
Down matrix of layer 20 in Llama2-7b.}

\textbf{(b) Bias towards minimum NSM in Llama2-7b: Attention Down matrix
of layer 25 in Llama2-7b.}

\textbf{(c) Bias towards maximum NSM in Zephyr: Attention Down matrix of
layer 9 in Zephyr.}

\textbf{(d) Bias towards minimum NSM in Zephyr: Attention Down matrix of
layer 20 in Zephyr.} There is only 6 minimum NSMs on the matrix of
Mmlu21 experiment.

\textbf{Figure 3: Patterns of NSM Enrichment in Rows and Columns}

\textbf{(a) Example of conserved row enrichment: Attention K matrix of
layer 5 across experiments in Llama2-7b.}

\textbf{(b) Conserved columnar enrichment: Attention V matrix of layer 3
in Llama2-7b.}

\textbf{(c) Multi-layer columnar enrichment: Conserved rows across
layers 0, 2, 3, 4, 5 of attention V matrix of python0 experiment in
Llama2-7b.}

\textbf{(d) Columnar enrichment in MLP Gate Matrix in Llama2-7b:
Specific to MMLU experiment in Llama2-7b (see Supplementary Figure 1 for
details).}

For ease of visualization, the images of all Up and Gates matrices are
transposed in this and following figures.

\textbf{Figure 4: Visualization of Correlative Complementary Pattern
(COPA) in Model Matrices}

\textbf{(a) COPA Visualization of attention matrices and MLP Down and Up
matrix in Llama2-7b:}

\textbf{Upper panels:} Samples of attention K Q V O matrix and MLP Down
and Up matrix from the Drosophila experiment, showing the initial
arrangement.

\textbf{Lower panels:} The same matrix rearranged by column to highlight
COPA.

\textbf{(b) COPA Visualization of attention matrices and MLP Down and Up
matrix in Zephyr:}

\textbf{Upper panels:} Samples of attention K Q V O matrix and MLP Down
and Up matrix from the Drosophila experiment, showing the initial
arrangement.

\textbf{Lower panels:} The same matrix rearranged by column to highlight
COPA.

\textbf{(c) COPA Visualization of MLP Gate matrices in Llama2-7b and
Zephyr:}

\textbf{Upper panels:} Samples of MLP Down matrix from the Java
experiment of Llama2-7b and Drosophila experiment of the Zephyr, showing
the initial arrangement.

\textbf{Middle panels:} The same matrix rearranged by row to highlight
COPA.

\textbf{Lower panels:} The same matrix rearranged by column to highlight
COPA.

\textbf{(d) Variance of Strength of COPA between MLP matrices of Python0
and Python 10 Experiment:}

General observation: Stronger complementary patterns in the Python10
experiment compared to the Python0 experiment. Matrices are rearranged
by column (Down matrix) or by row (Gate and Up matrix) for better
visualization of patterns.

Red arrows in lower panels indicate the reference vector used for
rearrangement. The rearrangement of only one axis is typically shown
because mathematically, a complementary pattern on one axis implies a
complementary pattern on the other axis. For clarity in figure
presentation, images of all Up and Gates matrices are transposed.

\textbf{Figure 5 Distribution and Enrichment of Malignant Mutations
Across Various Matrices and Experiments}

\textbf{(a) The conserved column 23 of the Attention V matrix in layer 1
of Llama2-7b enriches with underperforming NSMs and destructive NSMs in
the Mmlu21 experiment and mutations with low cosine similarity in the
Drosophila experiment.} Left panels include the Llama-7b Mmlu21 NSM maps
with thresholds set at various scores (2, 5, or 8), a specific map for
destructive NSMs, and an NSM map without threshold, showing that
underperforming NSMs with higher scores often appear in the same columns
as those with lower scores. Right panels include Drosophila NSM maps
with cosine similarity thresholds set at 0.1, 0.2, 0.5, or 0.7,
alongside an map without threshold, illustrating that NSMs with cosine
similarities below 0.6 tend to cluster within similar columnar
structures, regardless of their exact cosine similarity values.

\textbf{(b) Low cosine similarity mutations in the MLP Gate matrix of
layer 28 of Zephyr are enriched solely in the two rows that have
phenotypes for Mmlu21 experiment.} Left panels include the Zephyr Mmlu21
NSM maps with thresholds set at 2, 5, or 8, a dedicated map for
destructive NSMs, and an NSM map without threshold. These panels show
that underperforming NSMs from the MMLU21 experiment, irrespective of
their scores, consistently align within the same two rows. Right panels
include the Drosophila NSM maps with cosine similarity thresholds set at
0.1, 0.2, 0.5, or 0.7, and a map without threshold. NSMs from the
Drosophila experiment with cosine similarities below 0.5 also tend to
cluster within these same two rows.

\textbf{(c) The Drosophila, Newton, and p53 experiments of Zephyr
exhibit a highly conserved pattern in the distribution of mutations with
low cosine similarity in Down matrix of layer 30.} Upper Panels include
Zephyr NSM maps for the Drosophila, Newton, and P53 experiments with
specific cosine similarity thresholds set at 0.5, 0.6, and 0.6
respectively. These maps reveal a highly conserved columnar pattern of
mutations across these descriptive experiments---a rarity in other
matrices. Lower panels include NSM maps without threshold for the
Drosophila, Newton, P53, and MMLU21 experiments. Despite the pronounced
conservation of mutation patterns in the descriptive experiments, these
mutations do not impact performance in the MMLU21 task, as evidenced by
the entire Down matrix of layer 30 remaining silent (no NSMs) for the
MMLU21 experiment.

\textbf{(d) Severe malignant mutations in Llama2-7b that causing severe
disruptions for both the Drosophila experiment (cosine similarity
\textless{} 0.5) and the Mmlu21 experiment (score \textless=8)} Table
includes the number of malignant mutations that surpass the specified
thresholds. Pictures illustrate all matrices that contain at least 30
malignant NSMs.

\textbf{(e) Severe malignant mutations on Zephyr that causing severe
disruptions for both the Drosophila experiment (cosine similarity
\textless{} 0.5) and the Mmlu21 experiment (score \textless=8)} Table
includes the number of malignant mutations that surpass the specified
thresholds. Pictures illustrate all matrices that contain at least 30
malignant NSMs.

Most of the severe malignant mutations in D and E are also shared by the
other descriptive experiments (Supplementary Figure 5).

\textbf{Figure 6: Statistical Analysis and Distribution of RIHF NSMs and
Their Relation to Cosine Similarity}

\textbf{(a) Histogram of initial words of highest frequence:}

This histogram analyzes the initial words of highest frequency from 151
inputs across 136,522 NSMs for Zephyr and 127,350 NSMs for Llama2-7b
from the Drosophila experiments (See method). In Zephyr, 98\% of NSMs
commonly begin with "The," while the remainder feature a rare initial
word of highest frequency (RIHF) other than "The." In Llama2-7b, 98\%
start with "It" or "The," with the rest having RIHF different from these
words. The histogram displays the counts of NSMs in Zephyr (Left panel)
and Llama2-7b (Right panel), categorized by common or rare initial words
across different ranges of cosine similarity scores. NSMs with common
initial words typically exhibit higher cosine similarity values, while
those with RIHF are more evenly distributed among NSMs with a cosine
similarity of less than 0.8.

\textbf{(b) Distribution of all RIHF NSMs on different matrix Llama2-7b
and Zephyr.}

RIHF NSMs are primarily found in a small subset of attention and MLP
matrices, predominantly within the Down matrices.

\textbf{(c) Correlation of Coordinates for RIHF NSMs.}

Each plot visualizes the correlation for each RIHF between the number of
mutations (blue crosses) and the count of row(y) and column (x)
coordinates (red diamonds in left and right panel). The left set of red
diamonds represents the number of row coordinates, showing a significant
tendency for samples within a group to share row coordinates. In
contrast, the right set of red diamonds (number of column coordinates)
often matches the number of mutations, suggesting a random distribution
along the x-axis. This analysis only includes RIHFs with at least two
mutations.

\textbf{(d) Map of Low Cosine Similarity mutations and RIHF mutations of
Drosophila experiment on Down matrix of layer 1, 14, 30 of Zephyr.}

The diagram features red, light blue, and black arrows marking the
mutations of RIHF "staring," "otta," and "betrayal" in the Down matrix
of layers 1, 14, and 30. Mutations for each RIHF share one or several
row coordinates. Mutations listed in Supplementary Table 3 are
highlighted with a blue outline. A yellow arrow with a blue outline
indicates the mutation located in row G of Supplementary Table 3

\textbf{Supplementary Material}

https://github.com/anonymous-submission200/Mutagenesis-Screen-Of-LLM

\textbf{Methods}

\textbf{Coordinate Transformation:}

Each mutation affects a 64x64 square within the matrix. To facilitate
clarity and consistency in our visual presentations across the paper,
figures, and tables, we employ a coordinate transformation step.
Specifically, the coordinates are adjusted by dividing both the X and Y
values by 64. For example, if a mutation covers an area from (X, X+63)
to (Y, Y+63), the coordinates for this mutation are represented as
(X/64, Y/64).

\textbf{LLM Output Generation:}

Outputs were generated using the Hugging Face Transformers package for
both Llama2-7b and Zephyr. A temperature setting of 0.7 was utilized for
all experiments, except for the MMLU tasks. For the Python0, Python10,
Java, and Drosophila experiments, as well as the study of RIHF, the
`generate` function was set with a `max\_length` of 150. For the Newton
and P53 experiments, a `max\_length` of 300 was employed. Python10 used
a random seed of 10, while a seed of 0 was used for all other
experiments. For the MMLU task, the prompt with 5 in-context learning
examples from the chain-of-thought hub by Fu et al. {[}1{]} was used.
Specific prompts for the other experiments were as follows:

\begin{quote}
Python0 and Python10: The following is a python program for bubble sort:

Java: The following is a Java program for bubble sort:

Drosophila: The life cycle of Drosophila:

Newton: The scientific accomplishments and influences of Isaac Newton:

P53: Tell me 10 different signal pathways through which p53 is involved
in cancer development:
\end{quote}

\textbf{Mutation Overlap Calculation:}

For each 64x64 square within the matrix there are four possible statuses
depending on whether mutations were silent or non-silent mutations
(NSMs):

\begin{enumerate}
\def\labelenumi{\arabic{enumi}.}
\item
  Both maximum and minimum mutations of this square are NSMs.
\item
  Only the maximum mutation of this square is an NSM.
\item
  Only the minimum mutation of this square is an NSM.
\item
  Both mutations of this square are silent.
\end{enumerate}

The overlap ratio of mutations between two experiments, as illustrated
in Figure 1B, is determined by scoring each 64x64 square as follows:

\begin{itemize}
\item
  A score of 1 is assigned if the mutation status for both experiments
  within the square is identical.
\item
  A score of 0.5 is given if in one experiment, both maximum and minimum
  mutations are NSMs, and in the other, only one type (maximum or
  minimum) is an NSM.
\item
  A score of 0 applies to all other mismatched cases.
\end{itemize}

The total overlap ratio is calculated by summing the scores across all
squares and then dividing by the total number of squares covering the
matrix.

\textbf{Determining Matrix Bias Towards Maximum or Minimum NSMs}

To assess whether a matrix is biased towards maximum or minimum NSMs, we
utilize the following bias score formula:

\[bias\ score = \ \frac{|maximum\ NSM\ number - minimum\ NSM\ number|}{max(maximum\ only\ NSM\ number,minimum\ only\ NSM\ number)}\]

A matrix is considered biased towards maximum NSM if:

\begin{enumerate}
\def\labelenumi{\arabic{enumi}.}
\item
  The number of maximum NSMs exceeds the number of minimum NSMs.
\item
  The bias score exceeds 20\%.
\item
  The combined total of maximum only and minimum only NSMs is more than
  10\% of the total number of squares in the matrix.
\end{enumerate}

A matrix is considered biased towards minimum NSM if it satisfied above
criteria except it has more minimum NSMs than maximum NSMs.

\textbf{Bag-of-Words (BoW) Analysis and Cosine Similarity:}

Outputs were tokenized by splitting the text by punctuation marks
(period, comma, question mark, exclamation mark) and whitespace (newline
character, space). Tokens enclosed in brackets were stripped, and tokens
consisting solely of numbers were excluded. The tokens from all
different phenotypes of one experiment were pooled to generate a
bag-of-words vector space model. Each output was represented by a BoW
vector, with each entry representing the frequency of occurrence of a
distinct token {[}2{]}. Cosine similarity was then calculated between
the BoW vector of the standard output and that of the mutated
model\textquotesingle s output {[}2{]}.

\textbf{Initial Word of Highest Frequency Analysis:}

\textbf{Input Selection:} A total of 151 short sentences were used as
inputs for this analysis, comprising the specific input used in the
Drosophila experiment alongside 150 novel inputs generated by ChatGPT.
The composition of these inputs was deliberately structured to include
101 literary expressions and 50 factual statements.

\textbf{Output Generation and Frequency Calculation:} For both the
standard model and models with NSMs, outputs were generated using the
aforementioned inputs. The frequency of the initial words in these
outputs was then calculated, and the initial word of highest frequency
could be identified for both the standard model or model with specific
mutations.

\textbf{Mutation Selection for Analysis:} 136,522 NSMs for Zephyr and
127,350 NSMs for Llama2-7b were considered in this study. To ensure a
representative analysis, these mutations were selected to cover all the
unique phenotypes observed in the Drosophila experiment. For each unique
phenotype, no more than three mutations that produced this phenotype
were selected. Among these NSMs, there are 2192 NSMs for Zephyr with
Rare Initial Word of Highest Frequency (RIHF) that is not ``The'' and
1636 NSMs for Llama2-7b with RIHF that is not ``It'' or ``The''.

\textbf{Methods references}

{[}1{]} Fu Y, Ou L, Chen M, Wan Y, Pen H, Khot T ``Chain-of-Thought Hub:
A Continuous Effort to Measure Large Language Models' Reasoning
Performance'' In arXiv preprint arXiv:\textbf{2305}.17306, 2023.

{[}2{]} Salton G and McGill MJ ``\emph{Introduction to Modern
Information Retrieval}'' McGraw-Hill, Inc., New York, NY, USA, 1986.

\textbf{Competing interest declaration}

The authors declare that they have no conflict of interest.

\textbf{Additional information}

\end{document}